\documentclass{IEEEtran4PSCC}
\ifCLASSINFOpdf
   \usepackage[pdftex]{graphicx}
\else
   \usepackage[dvips]{graphicx}
\fi
\usepackage[cmex10]{amsmath}
\usepackage[T1]{fontenc}
\usepackage{url}
\usepackage[noadjust]{cite}
\makeatletter
\let\old@ps@headings\ps@headings
\let\old@ps@IEEEtitlepagestyle\ps@IEEEtitlepagestyle
\def\psccfooter#1{%
    \def\ps@headings{%
        \old@ps@headings%
        \def\@oddfoot{\strut\hfill#1\hfill\strut}%
        \def\@evenfoot{\strut\hfill#1\hfill\strut}%
    }%
    \def\ps@IEEEtitlepagestyle{%
        \old@ps@IEEEtitlepagestyle%
        \def\@oddfoot{\strut\hfill#1\hfill\strut}%
        \def\@evenfoot{\strut\hfill#1\hfill\strut}%
    }%
    \ps@headings%
}
\makeatother

\psccfooter{%
        \parbox{\textwidth}{\hrulefill \\ \small{23rd Power Systems Computation Conference} \hfill \begin{minipage}{0.2\textwidth}\centering \vspace*{4pt} \includegraphics[scale=0.06]{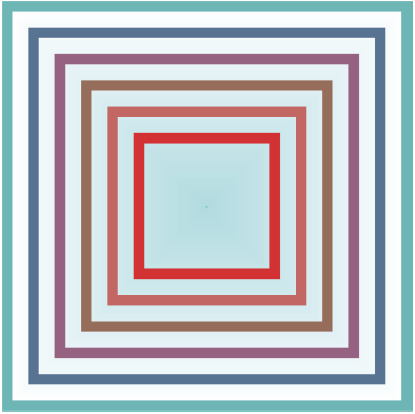}\\\small{PSCC 2024} \end{minipage} \hfill \small{Paris, France --- June 4 -- June 7, 2024}}%
}

\usepackage{algorithm}
\usepackage{algorithmicx}
\usepackage{graphicx}
\usepackage{xcolor}
\usepackage[caption=false,font=footnotesize]{subfig}

\usepackage{algcompatible}
\usepackage{algpseudocode}
\usepackage[cmex10]{amsmath}
\usepackage{amsfonts}
\usepackage{amssymb}
\usepackage{dsfont}
\usepackage{bm}
\usepackage{upgreek}
\usepackage{etoolbox}
\usepackage{multirow}
\usepackage{booktabs}
\usepackage{array}
\newcolumntype{R}{>{$}r<{$}}
\newcolumntype{L}{>{$}l<{$}}
\newcolumntype{M}{R@{${}/{}$}L}
\usepackage{siunitx}
\usepackage{hyperref}

\newcommand{\PS}{\mathcal{P}} 
\newcommand{\EDGES}{\mathcal{E}} 
\newcommand{\LOADS}{\mathcal{L}}
\newcommand{\GENERATORS}{\mathcal{G}}
\newcommand{\NODES}{\mathcal{N}}
\newcommand{\SG}{\mathbf{S}^{\text{g}}} 
\newcommand{\PG}{\mathbf{p}^{\text{g}}} 
\newcommand{\QG}{\mathbf{q}^{\text{g}}} 
\newcommand{\SD}{\mathbf{S}^{\text{d}}} 
\newcommand{\V}{\mathbf{V}}  
\newcommand{\VM}{\mathbf{v}} 
\newcommand{\VA}{\bm{\theta}} 
\newcommand{\SF}{\mathbf{S}^{\text{f}}} 

\newcommand{\bas}{\textsc{BAS}}
\newcommand{\acopf}{\textsc{ACOPF}}

\DeclareMathOperator*{\E}{\mathbb{E}}
\DeclareMathOperator*{\argmax}{\text{argmax}}
\DeclareMathOperator*{\argmin}{\text{argmin}}
\DeclareMathOperator{\supp}{\text{supp}}

\algrenewcommand{\algorithmiccomment}[1]{\Statex\hskip.25em ${\rhd}$ \texttt{#1}}

\newcommand{\sspace}{\vspace{1mm}}
\newcommand{\spacedhrule}{\sspace\hrule\sspace}

\newfloat{model}{thp}{lop}\floatname{model}{Model}

\begin{document}

\title{Bucketized Active Sampling for Learning ACOPF}

\author{
\IEEEauthorblockN{Michael Klamkin, Mathieu Tanneau, Terrence W.K. Mak, Pascal Van Hentenryck}
\IEEEauthorblockA{Industrial and Systems Engineering, Georgia Institute of Technology, Atlanta, GA, United States \\
\{klam, mathieu.tanneau, wmak, pascal.vanhentenryck\}@isye.gatech.edu}}

\maketitle

\begin{abstract}
  This paper considers optimization proxies for Optimal Power Flow (OPF),
  i.e., machine-learning models that approximate the input/output relationship of OPF.
  Recent work has focused on showing that such proxies can be of high fidelity.
  However, their training requires significant data, each instance necessitating the (offline) solving of an OPF.
  To meet the requirements of market-clearing applications, this paper proposes Bucketized Active Sampling (\bas{}), a novel active learning framework that aims at training the best possible OPF proxy within a time limit.
  \bas{} partitions the input domain into buckets and uses an acquisition function to determine where to sample next.
  By applying the same partitioning to the validation set, \bas{} leverages labeled validation samples in the selection of unlabeled samples.
  \bas{} also relies on an adaptive learning rate that increases and decreases over time.
  Experimental results demonstrate the benefits of \bas{}.
\end{abstract}

\begin{IEEEkeywords}
  ACOPF, machine learning, active learning
\end{IEEEkeywords}

\thanksto{\noindent This research was partly supported by NSF award 2112533 and ARPA-E PERFORM award AR0001136.}

\section{Introduction}\label{sec:introduction}
  The \emph{AC Optimal Power Flow} (\acopf{}) is a
  core building block in power systems
  that finds the most economical generation dispatch meeting the
  load demand, while satisfying the physical and operational
  constraints of the underlying systems.
  The problem is used in many applications including
  day-ahead security-constrained unit commitment ({SCUC})~\cite{sun2017decomposition},
  real time security-constrained economic dispatch ({SCED})~\cite{tam2011real},
  and expansion planning~\cite{verma2016transmission}.

  The non-convexity of \acopf{} limits the solving frequency of many operational
  tools. In practice, generation schedules in real time markets
  are required to be updated every 5 -- 15 minutes,
  varying from region to region.
  With the continuous integration of renewable energy sources
  and demand response mechanisms to meet renewable targets,
  load and generation uncertainties are expected to be more and more severe.
  Solving generation schedules using historical forecasts may no longer be
  optimal, and may not even be feasible in the worst case.
  Balancing generation and load rapidly without sacrificing
  economical efficiency is thus an important challenge.

  Recently, an interesting line of research has focused on how to learn
  and predict \acopf{} using optimization proxies, such as Deep Neural
  Networks (DNN)~\cite{fioretto2020predicting,yan20real,pan2020deepopf}.
  Once a DNN model is trained, predictions can be computed in
  milliseconds.  Preliminary results are encouraging, indicating that
  DNNs can predict \acopf{} solutions with high accuracy.  However, most
  prior work \cite{tang2017real,diehl2019warm,owerko2020optimal,dong2020smart,pan2020deepopf,zamzam2020learning,canyasse2017supervised,baker2020learning,guha2019machine} ignore the computational time
  spent gathering or sampling \acopf{} training data in practice. In particular, when
  generator commitments and grid topology change from day to day (or
  even within hours) e.g. to respond to wind forecast updates, it becomes
  impractical to develop a sophisticated learning mechanism that
  requires hours to sample and solve \acopf{} instances.  On the other
  hand, generating data sets and training a generalized framework
  ``once-for-all'' for all possible commitments, forecasts, and
  potential grid topologies is also unlikely to be feasible due to the
  exponential number of configurations \& scenarios of a transmission
  grid. Therefore, \textit{how to actively generate samples and train a
    ``Just-in-Time'' model} \cite{chen2021learning}, based on the
  current and forecasted information, is an important practical question
  for learning approaches.

  This paper is the first to study whether active sampling can address this important
  issue and, more generally, aims at decreasing the computational resources needed to
  train optimization proxies in practice. It proposes Bucketized Active Sampling (\bas{}), a novel active
  learning framework to perform ``Just-in-Time'' (JIT) training. Unlike
  many active learning frameworks that generate new samples based on the
  evaluations of individual (or groups of~\cite{kirsch2019batchbald})
  unlabeled data samples, \bas{} reasons about regions of the input
  domain, exploiting the fact that similar inputs should generally
  produce similar outputs. The input domain is thus bucketized and
  each bucket is evaluated (using an independent validation set) to
  determine if it needs additional samples.
  A variety of acquisition functions are examined, some
  inspired by several state-of-the-art approaches, e.g., BADGE \cite{ash2020deep},
  and MCDUE \cite{tsymbalov2018dropout}. \bas{} is evaluated on large transmission networks and
  compared to the state of the art in active sampling. The experimental
  results show that \bas{} produces significant benefits for
  JIT training.
  This paper's contributions can thus be summarized as follows:
  \begin{enumerate}

  \item The paper proposes \bas{}, a novel bucketized active learning
    framework for the Just-In-Time learning setting. Its key novelty is the idea of partitioning the input domain into buckets to determine which buckets are in need of additional samples. This bucket-based
    sampling contrasts with existing active learning schemes that
    determine which unlabeled samples to add based on the evaluation of
    their individual properties.

  \item To determine which buckets are in need of more samples, 
    \bas{} bucketizes the validation set in order to score each 
    \textit{bucket, thus}
    decoupling the assessment of
    where new samples are needed from the sample generation. This allows
    additional flexibility when learning optimization proxies, including
    the ability to avoid generating too many infeasible problems.

  \item The paper studies multiple acquisition functions for \bas{},
    highlighting the benefits of gradient information.

  \item The paper proposes a new dynamic adjustment scheme for the learning rate that
    may now decrease and {\it increase}.

  \item The paper presents an experimental evaluation of \bas{} on large
    \acopf{} benchmarks,
    demonstrating its benefits compared to state-of-the-art approaches.
  \end{enumerate}

  \noindent
  The rest of the paper is organized as follows. Section
  \ref{sec:background} presents background materials on \acopf{} and
  active sampling. Section \ref{sec:related} discusses related
  work on active sampling and \acopf{} learning.  Section
  \ref{sec:background:activesampling} introduces the novel active sampling framework.  Section \ref{sec:results} reports experimental evaluations and
  Section \ref{sec:conclusion} concludes the paper.

\section{Background}
  \label{sec:background}

    This paper uses calligraphic capitals $\mathcal{X}$ for sets and
    gothic capitals $\mathfrak{X}$ for distributions.
    The complex conjugate of $z \, {\in} \, \mathbb{C}$ is denoted $z^{\star}$.
    For decision variable $y$, its lower and upper bounds are denoted by $\overline{y}$ and $\underline{y}$, optimal value by $y^*$, and predicted value by $\hat{y}$.
    $\mathbf{x}$ and $\mathbf{y}$ represent the (flattened) vector of input and output features, respectively.

  \subsection{AC Optimal Power Flow}

    The AC Optimal Power Flow (\acopf{}) aims at finding the most economical
    generation dispatch to meet load demand, while satisfying both the
    transmission and operational constraints. The \acopf{} problem is nonlinear and
    non-convex.

    \begin{model}[!t]
      \caption{AC Optimal Power Flow --- \texttt{\acopf{}}($\SD$)}
      \label{model:acopf}
      \small
      \begin{subequations}
          \begin{align}
              \min_{\SG, \SF, \V} \quad
                  & \sum_{i \in \NODES} c_i (\SG_i) \label{model:acopf:obj} \\
              \text{s.t.} \quad
              & \SG_i - \SD_i = \sum_{(i,j) \in \mathcal{E}_{i} \cup \mathcal{E}^{R}_{i}} \SF_{ij}
                  & \forall i \in \NODES \label{model:acopf:kirchhoff} \\
              & \SF_{ij} = (Y_{ij} {+} Y^{c}_{ij})^{\star} \V_{i} \V_{i}^{\star} - Y_{ij}^{\star} \V_{i} \V_{j}^{\star}
                  & \forall (i,\, j) \in \EDGES \label{model:acopf:ohm:fr} \\
              & \SF_{ji} = (Y_{ij} {+} Y^{c}_{ji})^{\star} \V_{j} \V_{j}^{\star} - Y_{ij}^{\star} \V_{i}^{\star} \V_{j}
                  & \forall (i,\, j) \in \EDGES \label{model:acopf:ohm:to} \\
              & \underline{\VM_i} \leq |\V_i| \leq \overline{\VM_i}
                  & \forall i \in \NODES
                  \label{model:acopf:vmbound} \\
              & \underline{\SG_i} \leq \SG_i \leq \overline{\SG_i}
                  & \forall i \in \NODES
                  \label{model:acopf:genbound} \\
              & |\SF_{ij} |, |\SF_{ji}| \leq \overline{S_{ij}}^2
                  & \forall (i,\, j) \in \EDGES
                  \label{model:acopf:thrmbound}
          \end{align}
      \end{subequations}
    \end{model}
    
\begin{figure*}[t]
    \centering
    \subfloat[
        Traditional Active Sampling  (baseline).
        1) During active sampling, unlabeled candidate datapoints are sampled from a prescribed sampling distribution.
        2) Each candidate data point is scored according to a pre-defined metric; darker shades indicate higher scores.
        3) The $k$ data points with highest score are labeled and added to the training set.
        The new training data points are concentrated in the (top) right part of the input domain.
    ]{
        \label{fig:bas-tikz:baseline}
        \includegraphics[width=0.985\textwidth]{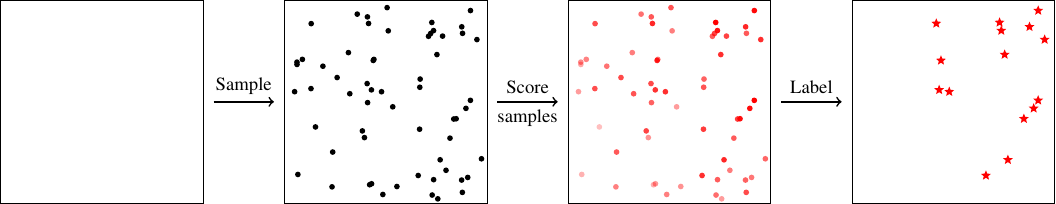}
    }\\
    \subfloat[
        Bucketized Active Sampling (proposed).
        1) Prior to training, the input domain is partitioned into buckets and a bucket validation set is sampled and labeled.
        2) During active sampling, each bucket is scored using the bucket validation set.
        3) $k$ data points are sampled and labeled; the higher a bucket's score, the more points are sampled from that bucket.
        The new training data points are spread across the input domain.
    ]{
        \label{fig:bas-tikz:BAS}
        \includegraphics[width=0.985\textwidth]{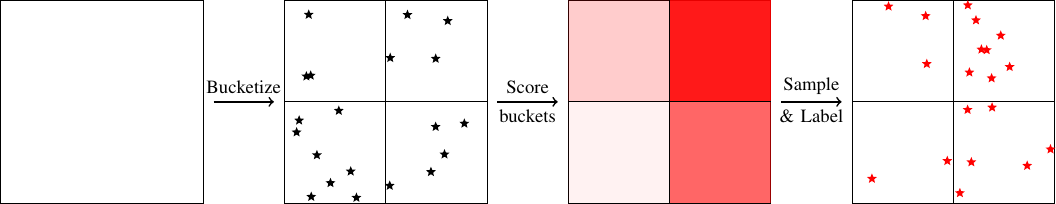}
    }\\
    \vspace{0.5em}
    \caption{Illustration of traditional active sampling (top) and the proposed bucketized active sampling (bottom) methodologies.
    The sampling distribution $\mathfrak{D}_{U}$ is a uniform distribution over the two-dimensional set $[0,1]^2$ (denoted by the square box).
    Legend: \textcolor{black}{$\bullet$} denotes an unlabeled data point; \textcolor{black}{$\boldsymbol{\star}$} denotes a labeled data point in the bucket validation set; \textcolor{red}{$\boldsymbol{\star}$} denotes a labeled data point that is added to the training set after active sampling.
    }
    \label{fig:bas-tikz}
\end{figure*}

    \begin{table}[t]
        \centering
        \caption{Nomenclature for Model~\ref{model:acopf}}
        \label{tab:nomen:acopf}
        \begin{tabular}{c|l}
          \toprule
          $\SG_i$  & Power generation at bus $i$  \\
          $\SF_{ij}$  & Power flow for branch $ij$           \\
          $\V_i$  & Voltage at bus $i$    \\
          \midrule
          $\SD_i$  & Power demand at bus $i$      \\
          $c_i$ & Cost per unit of generation at bus $i$ \\
          $\mathcal{E}_i$ & Set of branches originating at bus $i$ \\
          $\mathcal{E}^R_i$ & Set of branches terminating at bus $i$ \\
          $Y_{ij}$ & Line admittance for branch $ij$ \\
          $Y^c_{ij}$ & Shunt admittance for branch $ij$ \\
          $\overline{S_{ij}}$ & Thermal limit for branch $ij$ \\
          $\overline{\SG_i}$  & Generation upper bound at bus $i$   \\
          $\underline{\SG_i}$  & Generation lower bound at bus $i$  \\
          \bottomrule
        \end{tabular}
    \end{table}

    Let $\PS \, {=} \, \{\NODES,\, \LOADS,\, \GENERATORS,\, \EDGES\}$ be a power grid with buses $\NODES$, loads $\LOADS$, generators $\GENERATORS$, and transmission branches (lines and transformers) $\EDGES$.
    For each bus $i \, {\in} \, \NODES$, define $\EDGES_{i} \, {=} \, \{(i,\,j) \, {\in} \, \EDGES\}$ and $\EDGES^{R}_{i} \, {=} \, \{(i,\,j) \, {\in} \, \EDGES^{R}\}$
    where~$\EDGES^{R} \, {=} \, \{(j,\,i) \, | \, (i,\,j) \in \EDGES\}$. 
    For ease of presentation, the paper assumes that exactly one generator and one load is attached to each bus.
    The \acopf{} problem is formulated in Model~\ref{model:acopf}, where $\V$ denotes complex voltage, and $\SD, \SG, \SF$ denote complex power demand, generation and flow, respectively.
    The top and bottom sections of Table~\ref{tab:nomen:acopf} correspond to the decision variables and the problem data respectively.
    The line and shunt admittance of branch $(i,\,j)$ are denoted by $Y_{ij}$ and $Y^{c}_{ij}$.
    Objective \eqref{model:acopf:obj} minimizes total generation costs.
    Constraints \eqref{model:acopf:kirchhoff} maintain power balance (Kirchhoff's current law) for all the buses.
    Constraints \eqref{model:acopf:ohm:fr}, \eqref{model:acopf:ohm:to} assert Ohm's Law for all the lines.
    Constraint \eqref{model:acopf:vmbound} enforces upper and lower bounds for voltage magnitude $|\V|$.
    Constraints \eqref{model:acopf:genbound} impose bounds on active and reactive generation.
    Constraint \eqref{model:acopf:thrmbound} enforces the thermal limits $\overline{S_{ij}}$ on the apparent power for each line.
    For simplicity, Model~\ref{model:acopf} omits the detailed equations for bus shunts and transformers.
    The \acopf{} formulation used in the experiments follows the reference \acopf{} formulation in \texttt{PowerModels.jl} \cite{Coffrin2018_PowerModels} and its implementation; see also Model 1 in \cite{Coffrin2016_QCRelaxation}.

  \subsection{Dataset Generation for OPF proxies}

    Training optimization proxies for OPF requires generating a dataset of OPF instances and their solutions.
    This is done by sampling OPF instances according to a pre-specified sampling distribution $\mathfrak{D}_U$, then solving each sampled instance to obtain its solution.

    \begin{algorithm}[!b]
      \caption{\acopf{} Load Perturbation (\textsc{LoadPerturb})}\label{algo:acopfdbgen}
      \begin{algorithmic}[1]
        \State \textbf{Input:} Nominal sample $\SD_0$,
        regional load variation distribution $\mathfrak{B}$,
        individual noise variation distribution $\mathfrak{E}$, and data set size $n$.
        \State $\mathcal{D} \leftarrow \{\}$
        \For{$i=1\dots n$}
            \State $b_i \sim \mathfrak{B}$
            \State $\bm{\epsilon}_i \sim \mathfrak{E}$
            \State $\mathbf{x}_i \leftarrow b_i \cdot \SD_0 \circ \bm{\epsilon}_i$
            \State $\mathbf{y}_i~\leftarrow~(\SG,\,\V) \leftarrow\texttt{\acopf{}}(\mathbf{x}_i)$
            \If{$\mathbf{y}_i$ has been successfully found}
                \State $\mathcal{D} \leftarrow \mathcal{D} \cup \{(\mathbf{x}_i,\mathbf{y}_i)\}$
            \EndIf
        \EndFor
        \State \textbf{return } $\mathcal{D}$
      \end{algorithmic}
    \end{algorithm}

    Algorithm~\ref{algo:acopfdbgen} shows a random sampling procedure that
    perturbs load values from a nominal point $\mathbf{x}_0$.  The
    algorithm considers both regional influence and individual variations,
    modeled by the regional load variation distribution $\mathfrak{B}$ and the individual noise variation distribution
    $\mathfrak{E}$.  This general
    data generation pipeline is common among prior work, e.g.,
    \cite{pan2020deepopf,baker2020learning,guha2019machine,owerko2020optimal,dong2020smart,zamzam2020learning,canyasse2017supervised}, though
    not all prior approaches consider both regional and individual variations and
    not all prior approaches predict a complete \acopf{} solution, i.e., $(\SG, \V)$; note that $\SF$ are only used to express power flows.
    For instance \cite{guha2019machine} only predicts $\PG$ and some $\VM$, requiring a power flow to be solved to recover $\QG$, $\VA$, and the rest of $\VM$.
    {For ease of presentation, in the remainder of the paper, the distribution of \acopf{} instances defined by $\mathfrak{B}$ and $\mathfrak{E}$ is denoted by $\mathfrak{D}_U$.}
    In other words, in most prior work considering OPF proxies, the sampling distribution $\mathfrak{D}_U$ is parameterized by the distributions $\mathfrak{B}$ and $\mathfrak{E}$.

  \subsection{Active Sampling}

    Active sampling~\cite{angluin1988queries,lewis2017heterogeneous,kumar2020active,ren2021survey} aims at finding
    a small set $\mathcal{D}$ of data samples from a distribution
    $\mathfrak{D}_U$ to train a machine-learning model. In other words,
    given an unlabeled sample distribution $\mathfrak{D}_U$, the goal of
    active sampling is to design a query strategy  that chooses a small
    set of samples $\mathcal{D}$ from $\mathfrak{D}_U$ to label. Let
    $\mathbf{x}$ and $\mathbf{y}$ be the vector of input and output
    features of a data sample, i.e., $(\mathbf{x},\,\mathbf{y}) \in
    \mathcal{D}$, $h$ be a learning model, i.e., $h(\mathbf{x}) =
    \hat{\mathbf{y}}$, and $\mathbb{L}$ be a loss metric for measuring the
    prediction quality.  The data set $\mathcal{D}^*$ with the highest
    prediction accuracy is defined by a bilevel optimization problem:
    \begin{subequations}
        \label{eq:active_sampling}
        \begin{align}
        \mathcal{D}^* \in \argmin_{\mathcal{D}} & \enspace \E_{(\mathbf{x},\, \mathbf{y})\sim\mathfrak{D}} \left[ \mathbb{L}(h_{\mathcal{D}}^*(\mathbf{x}),\, \mathbf{y}) \right]\\
        \text{s.t. } 
            & \quad h_{\mathcal{D}}^{*} \in 
                \argmin_{h} \sum_{(\mathbf{x},\,\mathbf{y})\in\mathcal{D}} \mathbb{L}(h(\mathbf{x}),\, \mathbf{y})
        \end{align}
    \end{subequations}

    \noindent
    Solving \eqref{eq:active_sampling} optimally is challenging.  For many
    applications, drawing samples incrementally during training is more
    computationally efficient.  Let $\mathcal{D}^{S} =
    \supp(\mathfrak{D}_U)$ be the support set of distribution
    $\mathfrak{D}_U$.  Acquisition functions $\alpha$ (or scoring
    functions~\cite{choi2021active}) are commonly used in active sampling
    applications to select the best $n$ samples from $\mathcal{D}^{S}$ for
    the next training cycle.  The set $\mathcal{D}^*_{{n}}$ containing the $n$
    best samples drawn with acquisition function $\alpha$ is defined
    as:
    \begin{align} \label{eq:bmal}
        \mathcal{D}^*_{{n}} \in  \argmax_{\mathcal{D}_{{n}} \subseteq \mathcal{D}^S \mbox{ s.t. } |\mathcal{D}_{{n}}| = {{n}}} \enspace  \alpha\left(\mathcal{D}_{{n}}\right)
    \end{align}
    Note that acquisition functions are usually customized per
    application.
    Many acquisition functions analyze/query the learning
    model $h$ to select the next set $\mathcal{D}^*_{n}$ for training.

\section{Related Work}
  \label{sec:related}

  Most general active learning / sampling mechanisms
  \cite{wu2018pool,tsymbalov2018dropout,cai2013maximizing}
  label
  data samples based on
  acquisition scores of individual samples.
  Scores are often computed based on
  evaluations of the learning model, and
  the subset of samples with the highest scores will then be added to the data set.
  Some work also considers correlations between candidate input samples and/or
  ranking multiple \textit{unlabeled} samples at the same time as a batch, e.g.,  BatchBALD~\cite{kirsch2019batchbald}
  and BADGE~\cite{ash2020deep}.

  Contrary to prior approaches, \bas{} does not work at the level of
  individual
  samples: instead, it partitions the input domain into buckets and determines, based on a dedicated validation set,
  which buckets are in need of more samples.
  The validation set contains a list of buckets, where each
  bucket contains labeled validation samples for each individual partitioned subspace.
  \bas{} has several
  advantages: it exploits the structure of the underlying distribution,
  and it separates the analysis of the buckets from the sampling process.
  In the context of optimization proxies, \bas{} also makes it possible to
  avoid generating infeasible inputs.

  The effects of increasing learning rate during training have been the
  focus of several recent works including \cite{smith2017cyclical,smith2019super,loshchilov2016sgdr,zaheer2018adaptive}. However, prior active
  learning approaches focus only on decreasing the learning rate during
  training~\cite{liu2021influence,roy2018deep}. One of the novelties of
  the proposed scheme is to also {\em increase} the learning rate when appropriate, avoiding premature termination.

  \section{Active Learning for \acopf{}} \label{sec:background:activesampling}

    This section describes the novel active learning framework \bas{}.
    Figure \ref{fig:bas-tikz} illustrates a baseline active sampling and the proposed \bas{} scheme on a simple, two-dimensional example.
    As can be seen in Figure \ref{fig:bas-tikz:baseline}, in the baseline active sampling approach, most new training data points are added in the top-right region of the input domain.
    In contrast, as illustrated in Figure \ref{fig:bas-tikz:BAS}, \bas{} naturally distributes new samples across the input domain.
    This ensures that (i) new samples are not added in a single region of the input space, and (ii) computational resources are not wasted if that region turns out to contain mainly infeasible samples.
    
    The rest of the section first presents the Deep Neural Network (DNN) for \acopf{} learning. It
    then introduces the core idea of \bas{} before going into the
    individual components in detail.

    \begin{algorithm}[!t]
      \caption{\textsc{Active Learning Loop}}
      \label{algo:main}
      \begin{algorithmic}[1]
        \Statex \textbf{Parameters:}
        Initial learning rate $\beta_0$; Learning rate bounds $\overline{\beta}$ \& $\underline{\beta}$;
        Patience thresholds $\overline{\rho_1}$ \& $\overline{\rho_2}$;
        Learning rate factors $\gamma_1$ \& $\gamma_2$; Termination condition
        $\textsc{ShouldStop}(\cdot)$
        \Statex \textbf{Inputs:} Initial training data set $\mathcal{D}_0$; Bucket-validation set $\mathcal{D}_V$;
        DNN $h_\theta$
        \Statex \textbf{Outputs:} DNN $h_\theta$
        \spacedhrule
        \State $\beta\leftarrow\beta_0$, $l^*_V \leftarrow \infty$, $\rho_1 \leftarrow 0,  \rho_2\leftarrow0$, $\mathcal{D} \leftarrow \mathcal{D}_0$, \label{algo:main:init}
        \While{$\neg\textsc{ShouldStop}(\cdot)$}  \label{algo:main:while}
          \sspace
          \Comment{Training}
          \State $h_\theta \leftarrow \textsc{Train}(h_\theta,\, \mathcal{D},\, \beta)$ \label{algo:main:train}
          \sspace
          \sspace
          \Comment{Validation \& Patience Update}
          \State $l_V \leftarrow \displaystyle\sum_{(\mathbf{x}, \mathbf{y}) \in \mathcal{D}_V} \mathbb{L}(h_\theta(\mathbf{x}),\, \mathbf{y})$ \label{algo:main:val}
          \sspace
          \State $\rho_1,\, \rho_2,\, l^*_V,\, \beta \leftarrow{\textsc{Patience}}(\rho_1,\, \rho_2,\, l_V,\, l^*_V,\,\beta)$ \label{algo:main:patiencestep}
          \sspace
          \sspace
          \Comment{Active Sampling}

          \If{$\rho_2=0$} \label{algo:main:samplecond}
            \State $\mathcal{D} \leftarrow \mathcal{D} \cup {\bas{}}\left(h_\theta,\mathcal{D}_V\right)$ \label{algo:main:sample}
          \EndIf \label{algo:main:samplecondclose}
        \EndWhile \label{algo:main:end}
      \end{algorithmic}
    \end{algorithm}

    \begin{algorithm}[!t]
    \caption{\textsc{BAS: Bucketized Active Sampling}}\label{algo:activesample}
      \begin{algorithmic}[1]
        \Statex \textbf{Parameters:}
        Acquisition function $\alpha$;
        Distributor function $\eta$; Maximum number of new samples $\overline{n}$
        \Statex {
        \textbf{Inputs:} DNN $h_\theta$, Bucket-validation set $\mathcal{D}_V$}
        \Statex {\textbf{Output:} Set of new samples $\mathcal{D}^{+}$}
        \spacedhrule
        \State \textbf{Partition} bucket-validation data into a set of $k$ buckets \label{algo:activesample:partition}
          \Statex \quad $\{B_1, B_2, \dotsc, B_k\} \leftarrow \textsc{Partition}(\mathcal{D}_V)$
        \State \textbf{Evaluate} each bucket with acquisition function $\alpha(\cdot)$ \label{algo:activesample:evaluate}
          \Statex \quad $\mathcal{S} = \{ s_i : s_i \leftarrow \alpha(B_i, h_\theta), \forall i \in [1,k]\}$
        \State \textbf{Convert} scores $\mathcal{S}$ to the number of samples to be drawn from each bucket with distributor function $\eta(\cdot)$ \label{algo:activesample:convert}
          \Statex \quad $\mathcal{N} = \{n_1, n_2, \dotsc, n_k\} \leftarrow \eta(\mathcal{S},\overline{n})$
        \State {\textbf{Sample} $n_i$ new inputs from each bucket $B_i$} \label{algo:activesample:sample}
          \Statex \quad $\mathcal{X}_i^+ \leftarrow
        \left\{\mathbf{x}_j : \mathbf{x}_j \sim\mathfrak{D}_{B_i}, \forall j \in [1, n_i]\right\}, \forall i \in [1,k]$
        \State {\textbf{Compute} \acopf{} solutions for new samples $X_i^+$} \label{algo:activesample:compute}
          \Statex \quad ${\mathcal{Y}_i^+ \leftarrow \{\mathbf{y}_j :
        \mathbf{y}_j \leftarrow \texttt{\acopf{}}(x_j), \forall j \in [1, n_i]
        \}},$
          \Statex $ \quad\quad\quad\quad \forall i \in [1,k]$
        \State {\textbf{Return} $\mathcal{D}^+$} \label{algo:activesample:return}
          \Statex \quad $\mathcal{D}^+_i = \{(\mathbf{x}_j, \mathbf{y}_j) : \mathbf{x}_j \in \mathcal{X}_i^+, \mathbf{y}_j \in \mathcal{Y}_i^+, \forall j \in [1,n_i]  \}$
          \Statex $ \quad\quad\quad\quad \forall i \in [1,k]$
          \Statex \quad $\mathcal{D}^+ \leftarrow \bigcup_{i \in [1,k]} \mathcal{D}^+_i$
      \end{algorithmic}
    \end{algorithm}

  \subsection{Active Learning: DNN Model}

    The \acopf{} problem in Model~\ref{model:acopf} can be viewed as a
    high dimensional nonlinear function $h_{\mbox{ac}}$, taking input
    vector $\SD$, and returning optimal values for $\SG$ and $\V$. The learning task for OPF typically
    generates a finite set $\mathcal{D}$ of data samples, and each sample
    $(\mathbf{x},\,\mathbf{y}) \in \mathcal{D}$ is of the form:
    $\mathbf{x} = (\SD), \enspace \mathbf{y} = (\SG,\,\V).$
    Subscripts on $\mathbf{x}/\mathbf{y}$ identify particular samples,
    e.g., $\mathbf{x}_i$ refers to the input features of the
    $i^{\text{th}}$ sample.

    This work uses Deep Neural Networks (DNNs) to illustrate the benefits
    of active learning for speeding up \acopf{} learning. A DNN $h_\theta$ is
    composed of a sequence of $t$ layers, with each layer taking as input the
    results of the previous layer~\cite{lecun2015deep}:
    \begin{align*}
      \mathbf{h}^0(\mathbf{x}) &= \mathbf{x} & \\
      \mathbf{h}^j(\mathbf{x}) &= \pi(\mathbf{W}^j \mathbf{h}^{j - 1}(\mathbf{x}) + \mathbf{b}^j) &\forall j \in [1,t] 
    \end{align*}
    where
    $\mathbf{x}$,
    $\mathbf{W}^j$,
    $\mathbf{b}^j$,
    and $\pi$
    define the input vector, the $j^\text{th}$ weight matrix, the
    $j^\text{th}$ bias vector, and the activation function respectively. $\theta$ refers to the set of all $\mathbf{W}^j$ and $\mathbf{b}^j$ for $j\in[1, t]$.
    The problem of training a DNN $h_\theta$ on a dataset $\mathcal{D}$ consists of finding the parameters $\theta^*$ that minimize empirical risk:
   \begin{align}
       \theta^* \in \argmin_{\theta} \sum_{(\mathbf{x},\,\mathbf{y})\in\mathcal{D}} \mathbb{L}(h_\theta(\mathbf{x}),\,  \mathbf{y}) \label{eq:erm}
   \end{align}

\subsection{The Active Learning Scheme for BAS}

    Algorithm~\ref{algo:main} depicts the overall structure of the training loop. The
    main contributions are in the algorithm for active sampling presented in Section \ref{sec:bas} and Algorithm~\ref{algo:activesample}.

    The active sampling algorithm (Algorithm 2) receives an initial training data set $\mathcal{D}_0$, a
    validation data set $\mathcal{D}_V$, and an initialized DNN
    model $h_\theta$ as inputs.  The initial and validation data sets
    $\mathcal{D}_0$ and $\mathcal{D}_V$ can be constructed from
    Algorithm~\ref{algo:acopfdbgen} or
    from historically known \acopf{} solutions.
    Let $\mathfrak{D}_{U}$ be the sampling distribution 
    from which the inputs are drawn.
    Prior to training, the input domain is partitioned into buckets according to some task-specific domain knowledge.
    Any finite partitioning of the input domain is valid for \bas{}, i.e., each bucket should be non-empty and the input domain should be covered by the union of all buckets.
    For the \acopf{} problem at hand, the paper uses a partitioning strategy based on the one-dimensional base load factor $b_i\sim\mathfrak{B}$; see Section \ref{sec:bucketvalidation} for details.
    Note that general partitioning schemes such as $k$-medoids may also be used.
    This partitioning is applied to both $\mathfrak{D}$ 
    and $\mathcal{D}_V$ to form buckets containing 
    a (potentially infinite) set of unlabeled samples 
    from $\mathfrak{D}$ and a finite set of corresponding 
    labeled validation samples from $\mathcal{D}_V$. 
    Note that the partitioning scheme should be such
    that it allows for sampling from a particular 
    partition/bucket. Line 1 initializes the routine
    variables including the learning rate, best validation loss, patience
    counters, and training set.
    Lines \ref{algo:main:while}-\ref{algo:main:end} constitute
    the main loop of the training algorithm, which terminates when $\textsc{ShouldStop}(\cdot)$ is true, typically a bound on time or iterations.
    Line \ref{algo:main:train} updates the parameters of the DNN $h_\theta$ using the dataset $\mathcal{D}$ and learning rate $\beta$.
    Line \ref{algo:main:val}
    computes the validation loss using the bucket-validation
    set. Line \ref{algo:main:patiencestep} calls the $\textsc{Patience}$ routine
    (Algorithm~\ref{algo:patience}) to update the patience counters
    $\rho_1,\, \rho_2$ and the learning rate $\beta$. See Section \ref{sec:patience} for more details on patience-based scheduling. Finally, if the patience counter $\rho_2$ exceeds its threshold, new samples
    are drawn and their \acopf{} is computed (Lines \ref{algo:main:samplecond} - \ref{algo:main:samplecondclose}) following \bas{} (Algorithm~\ref{algo:activesample}).

  \subsection{Bucketized Active Sampling}
    \label{sec:bas}
    Algorithm~\ref{algo:activesample} presents the active sampling
    routine \bas{}.
    The key innovation of \bas{} is that by partitioning
    the input domain into buckets
    that capture the inherent structure of
    the application at hand,
    \bas{} can decouple the assessment of where
    additional samples are needed from the sample generation.
    Instead of generating samples based on their
    \textit{individual} qualities, \bas{} uses 
    a pre-chosen set (the bucket-validation set) 
    to determine the \textit{buckets} where samples 
    are most needed. By comparing properties across
    different buckets, \bas{} effectively allocates
    computational resources to generate more samples in poorly performing buckets,
    resulting in a more efficient sampling and learning routine.

  \begin{figure*}
    \centering
    \includegraphics[width=\textwidth]{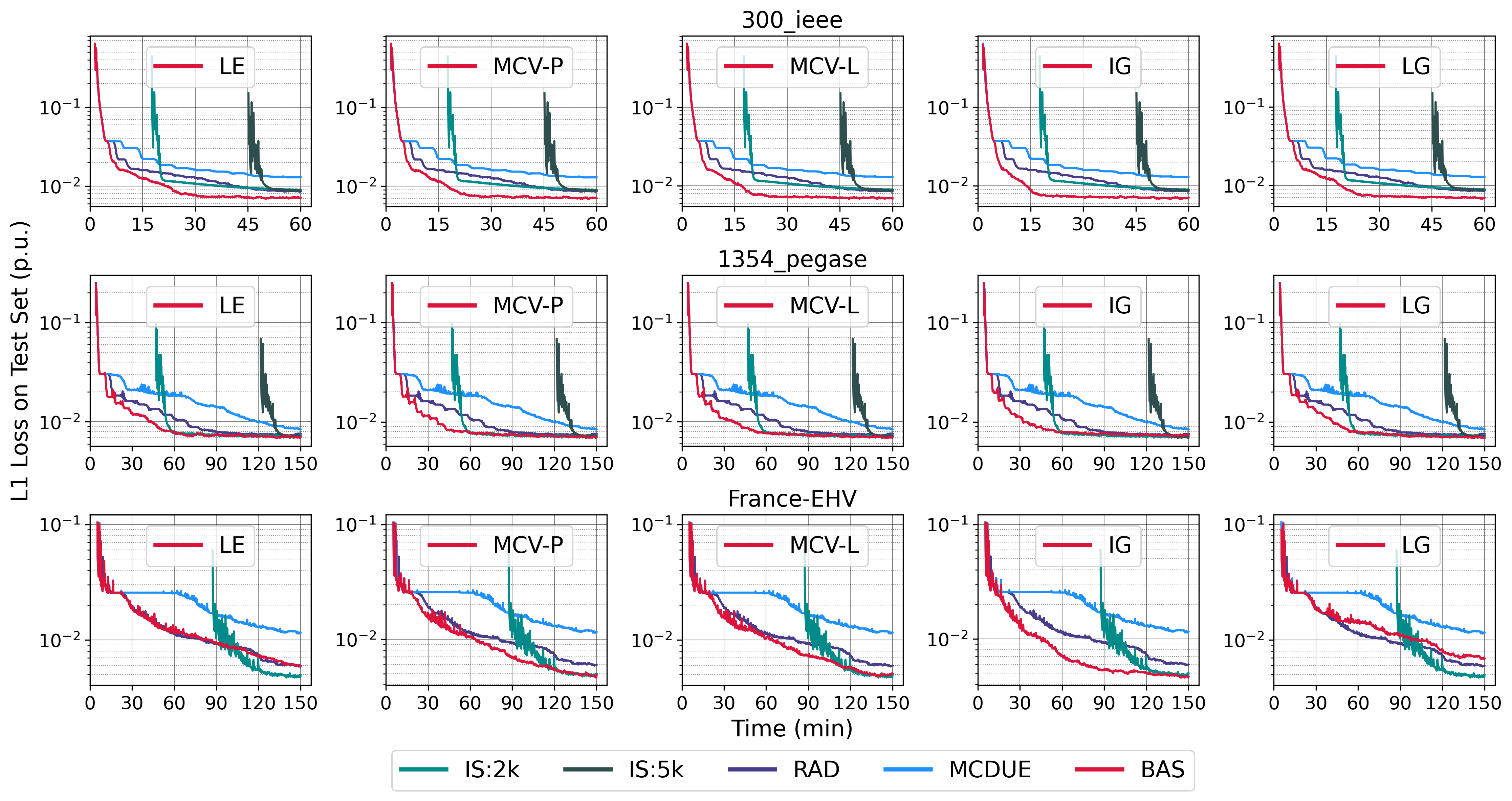}
    \caption{Mean L1 Loss on testing set, averaged across 50 trials.}
    \label{fig:bigloss}
  \end{figure*}
    
  The general process is described next.
  Step~\ref{algo:activesample:partition} partitions the bucket-validation data set. Step~\ref{algo:activesample:evaluate} evaluates the buckets using the acquisition function $\alpha$ and returns a set $\mathcal{S}$ of scores $s_i$, reflecting how urgently new samples are needed in bucket~$i$. Step~\ref{algo:activesample:convert} uses a distributor function to decide how many samples to generate for each bucket based on the scores. Step~\ref{algo:activesample:sample} generates the input feature vectors $\mathbf{x}_j$ by sampling from a predefined distribution $\mathfrak{D}_{B_i}$ for each bucket $i$. See \ref{sec:bucketvalidation} for how $\mathfrak{D}_{B_i}$ can be defined.
  Step~\ref{algo:activesample:compute} computes the \acopf{} solutions and records the output $\mathbf{y}_j$ for each generated input $\mathbf{x}_j$ across all the buckets.
  Step~\ref{algo:activesample:return} reconstructs the new samples and returns them to the main learning routine.

    Step 2 of \bas{} requires an acquisition function $\alpha$ to  score each bucket $B_i$. In \bas{}, $\alpha$ is defined in terms of $\mathbb{M}$:
    \begin{align}
      \label{eq:alpha}
      \alpha(B_i, h_\theta) & \equiv  \frac{1}{\lvert B_i \rvert} \sum_{(\mathbf{x}_j, \mathbf{y}_j) \in B_i} \mathbb{M}(h_\theta, \mathbf{x}_j, \mathbf{y}_j)
    \end{align}
    where $\mathbb{M}$ is a function measuring the score of individual
    samples that has access to the weights \& bias values of the training
    model, and every individual sample (and its label) in the bucket-validation set.
    \bas{} is evaluated on the following choices of acquisition metric $\mathbb{M}$:

    \begin{enumerate}
      \item Loss Error (LE): \hfill
      $\mathbb{L}(h_\theta(\mathbf{x}_j), \mathbf{y}_j)$
      \sspace\sspace
      \item Input Loss Gradient (IG): \hfill
      $\lVert \nabla_{\mathbf{x}_j} \mathbb{L}(h_\theta(\mathbf{x}_j), \mathbf{y}_j) \rVert$
      \sspace\sspace
      \item Last-layer Loss Gradient (LG): \hfill
        $\lVert \nabla_{\mathbf{W}^{t}} \mathbb{L}(h_\theta(\mathbf{x}_j), \mathbf{y}_j) \rVert$
        \sspace\sspace
        \item Monte-Carlo Prediction Variance (MCV-P), where each $h_\theta^i, i \in [1..\tau]$ has a random dropout mask:
        $$\mathrm{Var}\{h_\theta^1(\mathbf{x}_j), h_\theta^2(\mathbf{x}_j), \dotsc h_\theta^\tau(\mathbf{x}_j) \}$$

      \item Monte-Carlo Loss Variance (MCV-L):
        $$\mathrm{Var}\{\mathbb{L}(h_\theta^1(\mathbf{x}_j), \mathbf{y}_j), \mathbb{L}(h_\theta^2(\mathbf{x}_j), \mathbf{y}_j), \dotsc \mathbb{L}(h_\theta^\tau(\mathbf{x}_j), \mathbf{y}_j) \}$$
    \end{enumerate}

    MCV-P and LG resemble the acquisition functions proposed in
    \cite{tsymbalov2018dropout} and \cite{ash2020deep} respectively, 
    though in BAS they are computed using \textit{labeled samples}
    from the bucket-validation set rather than an \textit{unlabeled 
    pool}, allowing BAS to exclude infeasible samples from metric 
    calculation.

    Step 3 of \bas{} requires a distributor function $\eta$ to decide how
    to distribute the samples based on the scores computed by $\alpha$.
    The framework requires $\eta$ to return a set $\mathcal{N}$ satisfying \mbox{$0 \leq
    \sum_{n_i \in \mathcal{N}} n_i \leq \overline{n}$}. \bas{} is evaluated with a simple proportional distributor:
        $$
          \mathcal{N} = \{\overline{n}\frac{ s_i}{\sum_{s_j \in S} s_j}\,\vert \,\forall s_i \in S\}
        $$
    \begin{algorithm}[!t]
      \caption{\textsc{Patience}$(\rho_1, \rho_2, l_V, l^*_V, \beta)$}\label{algo:patience}
      \begin{algorithmic}[1]
        \If{$l_V  \geq   l_V^*$}
          \State $\rho_1 \leftarrow \rho_1 + 1$, $\rho_2 \leftarrow \rho_2 + 1$
        \Else
          \State $l_V^*\leftarrow l_V$, $\rho_1 \leftarrow 0$
        \EndIf
        \If{$\rho_1>\overline{\rho_1}$}
          \State $\beta\leftarrow\min\{\max\{\gamma_1\beta,\enspace \underline{\beta}\},\enspace \overline{\beta}\}$
          , $\rho_1 \leftarrow 0$
        \EndIf
        \If{$\rho_2>\overline{\rho_2}$}
          \State $\beta\leftarrow\min\{\max\{\gamma_2\beta,\enspace \underline{\beta}\},\enspace \overline{\beta}\}$
          , $\rho_2 \leftarrow 0$
        \EndIf
        \State {\textbf{return:} $\rho_1, \rho_2, l^*_V, \beta$}
      \end{algorithmic}
    \end{algorithm}
  \subsection{Patience-based Scheduling}
    \label{sec:patience}
    
  Another novelty of \bas{} is its use of patience-based scheduling, inspired by 
  \mbox{\textsc{ReduceLROnPlateau}}~\cite{pytorch}.
  \bas{} uses two patience
  counters $\rho_1$ and $\rho_2$. Counter $\rho_1$ is paired with
  $\gamma_1$, the classical discount factor ($\gamma_1 < 1.0$), to
  reduce the learning rate $\beta$ when $\rho_1$ reaches its threshold
  $\overline{\rho_1}$. When $\rho_2$ reaches its threshold, active sampling is triggered. In addition, to avoid premature learning
  convergence when new samples are added, \bas{} uses a boosting factor
  $\gamma_2> 1.0$ to increase the learning rate. Algorithm~\ref{algo:patience} describes the procedure.

\section{Experimental Evaluation}\label{sec:results}

  Experiments are performed on
    8 Intel Xeon 6226 CPU (2.7 GHz) machines, each with 192GB RAM.
    A Tesla V100-32GB GPU is used for training DNNs.
    Nonlinear solver Ipopt \cite{biegler2009large} with linear solver MA57 \cite{duff2004ma57} is used to compute \acopf{} solution data sets,
    and active learning experiments are implemented in Python 3.9.12 \cite{python} with
    PyTorch 1.11.0~\cite{pytorch}.

  \subsection{Benchmarks}

    The experimental results are presented for public benchmark power
    networks 300\_ieee and 1354\_pegase from
    PGLib \cite{babaeinejadsarookolaee201_PGLib_short} as well as France-EHV, an industrial proprietary benchmark with 1737 buses, 1731 loads, 290 generators, and 2350 lines/transformers based on the French EHV grid.

  \subsection{Bucket-Validation Set Construction}\label{sec:bucketvalidation}
    The evaluation uses Algorithm~\ref{algo:acopfdbgen} to
    generate datasets with \mbox{$\mathfrak{B}=\text{Uniform}(0.8, 1.2)$} and
    $\mathfrak{E}=\text{LogNormal}(0, 0.05)$. For simplicity, $\mathfrak{D}_{B_i}$ is defined as a projection from the original sampling distribution $\mathfrak{D}_U$ onto the domain/feasible space of each bucket $B_i$. In other words, each bucket $B_i$ contains samples $\mathbf{x}_j$ whose load factor $b_j$ is within the  lower/upper bounds $\underline{b_i}/\overline{b_i}$: 
    $\left(\mathbf{x}_j \sim \mathfrak{D}_{B_i} \right) \equiv \left(\mathbf{x}_j \sim \mathfrak{D} \mbox{ where } b_j \in [\underline{b_i}, \overline{b_{i}}]\right)$.
    All buckets of the bucket-validation set
    have $\lfloor\frac{n_V}{k}\rfloor$ feasible samples except for the last bucket which may have less. The evaluation uses $n_V=1024$ and $k=30$.

  \subsection{Learning Models}

    All experiments use fully connected DNNs with Sigmoid activation functions.
    Hidden layer dimensions are set to match
    the prediction feature size $2\vert\GENERATORS\vert +
    \vert\NODES\vert+\vert\EDGES\vert$.
    1354\_pegase and France-EHV experiments
    use 5 hidden layers while
    300\_ieee experiments use 3.
    The training uses a mini-batch
    size of 128, the AdamW optimizer
    \cite{loshchilov2017decoupled}, and the L$_1$ norm as the loss
    function. 256 samples are
    labeled to create an initial training set and 5000 testing samples are used for
    evaluations.
    Models are trained for at most 150 minutes, including training set solve time.
    Results are reported on held-out test data and are averaged over 50 trials using different random seeds.

    \bas{} is compared with three baseline methods:
    \begin{enumerate}
      \item Inactive Sampling (IS): training
        data is pre-generated by sampling
        uniformly from the input domain.

      \item Random Active Sampling (RAD): an active method that
        selects samples uniformly from the input domain.

      \item Monte-Carlo Dropout Uncertainty Estimation (MCDUE):
        the state-of-the-art active method from~\cite{tsymbalov2018dropout}.
    \end{enumerate}

    \noindent
    Inactive sampling is used to show the benefits of sampling data
    actively.  MCDUE is a state-of-the-art method in active learning, with
    the number of trails $\tau = 25$ as in~\cite{tsymbalov2018dropout}.
    At every iteration, a pool of 5000 unlabeled samples is generated
    uniformly from the input domain for MCDUE to select from.

    \begin{figure*}
      \centering
      \includegraphics[width=\textwidth]{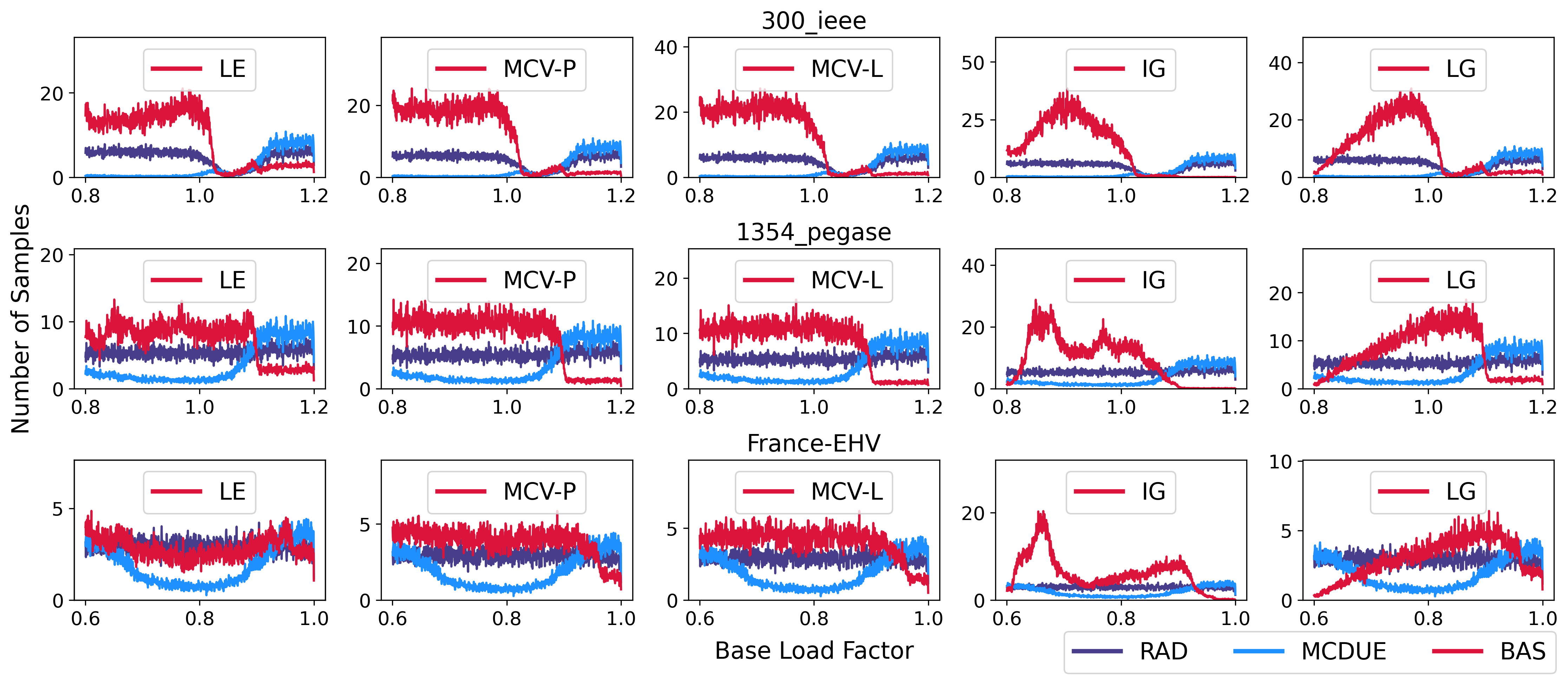}
      \caption{Mean training set distributions at termination.}
      \label{fig:sampledist}
    \end{figure*}

  \subsection{Learning Quality: Accuracy \& Convergence}

    Figure \ref{fig:bigloss} presents the prediction accuracy (L1-loss)
    over the training period.    Each row corresponds to
    a different benchmark, and each column corresponds to a different
    acquisition function.
    Overall, the prediction accuracy of \bas{} is
    consistently better than the other approaches.  Both RAD and MCDUE
    take significantly more time to match the level of accuracy of \bas{}.
    Even though classic generate-then-train (IS) approaches may converge
    faster, they suffer from the significant amount of time needed to
    pre-generate data samples. Observe also that \bas{}-IG consistently has the lowest prediction errors
    over the training horizon. Interestingly, MCV-P, MCV-L, and IG
    converge faster than LE and LG.

  \subsection{Sampling Behavior of Acquisition Functions}

    Figure \ref{fig:sampledist} shows the distribution of samples,
    explaining why \bas{}-IG is most effective. Observe the fundamentally
    different sampling behavior of IG compared to the other acquisition
    functions. LE, MCV-P, and MCV-L have no clear preferences across
    buckets (except in the higher ranges of base loads, which will be
    discussed shortly), reducing the benefits of active learning. LG has a
    clear bias towards samples with higher base loads, as it uses the
    gradient of the loss with respect to the last layer.  In contrast, IG
    has a much more differentiated sampling pattern. Because it uses the
    gradient of loss with respect to the input features, {\em IG isolates
      highly sensitive regions of the input space}.  These samples trigger
    significant weight updates, even after passing through \emph{all the
      DNN layers} during back-propagation.  Thus, IG captures information
    from both forward evaluations and backward propagations. LE, MCV-P,
    and MCV-L mostly capture forward-information (prediction and loss),
    and fail to differentiate the different regions. LG has
    too large a bias towards samples with large outputs.

    Second, observe that all acquisition functions under-sample regions
    with high load factor (e.g., $>$1.1 for
    1354\_pegase), contrary to MCDUE. These regions contain many
    infeasible samples which are not realistic as they correspond to
    loads that are over either global or regional production capabilities.  Uncertainty-based active sampling methods (e.g., MCDUE)
    select samples with the highest predictive variance, which often
    correspond to infeasible samples.  In contrast, the
    bucket-validation set allows \bas{} to allocate computational
    resources efficiently by surgically sampling in feasibility-dense
    buckets, which are more useful for learning. Table
    \ref{table:sample-numbers} highlights these observations further 
    by reporting the number of feasible and infeasible samples generated by each active sampling method
    within the time limit,
    where the top section shows the RAD and MCDUE baselines
    and the bottom section shows \bas{} with various acquisition functions.
    The table
    shows that \bas{}-IG is also able to label many more samples within
    the allocated time, limiting the number of time-consuming infeasible samples
    and focusing on regions of the input space that are more sensitive but
    whose samples are typically cheaper to label.

    \begin{table}[!t]
      \centering
      \caption{Mean number of feasible / infeasible samples generated}
      \label{table:sample-numbers}
          \begin{tabular}{l|MMM}
            \toprule
              Method & \multicolumn{2}{c}{\makebox[0pt]{300\_ieee}} & \multicolumn{2}{c}{\makebox[0pt]{1354\_pegase}} & \multicolumn{2}{c}{\makebox[0pt]{France-EHV}} \\
            \midrule
              {RAD} & 3275 & 2613 & 3937 & 1477 & 2651 & 266 \\
              {MCDUE} & 290 & 3462 & 1455 & 2078 & 1554 & 334 \\
            \midrule
              {BAS:LE} & 8019 & 1406 & 6306 & 887 & 2556 & 274 \\
              {BAS:MCV-P} & 10070 & 1000 & 7519 & 533 & 3621 & 214 \\
              {BAS:MCV-L} & 10818 & 883 & 7723 & 483 & 3794 & 203 \\
              {BAS:IG} & \textbf{11396} & \textbf{766} & \textbf{8860} & \textbf{135} & \textbf{5802} & \textbf{104} \\
              {BAS:LG} & 8967 & 1129 & 6709 & 758 & 2718 & 264 \\
            \bottomrule
          \end{tabular}
      \end{table}

\section{Conclusion \& Future Work} \label{sec:conclusion}

This paper proposed \bas{}, a novel bucketized active learning
framework. The framework partitions the input domain into
buckets and labels data samples based on the bucketized input space.
A key innovation of \bas{} is the use of a bucket-validation set to
decide where new samples are needed, decoupling the decision about
where to sample from the actual sample generation. \bas{} also adopts
a new learning rate adjustment scheme, and incorporates multiple
acquisition functions inspired by the literature.  Experimental
evaluation
indicates that \bas{} converges faster than state-of-the-art approaches
on large \acopf{} benchmarks.  Future work to
improve \bas{} includes dynamic schemes for the selection of
validation sets and bucket partitioning, and
cost-aware acquisition functions.

\bibliographystyle{IEEEtran}
\bibliography{refs}

\end{document}